\documentclass[journal]{IEEEtai}

\usepackage{amsmath,amsfonts}
\usepackage{algorithm}
\usepackage[caption=false,font=normalsize,labelfont=sf,textfont=sf]{subfig}
\usepackage{textcomp}
\usepackage{stfloats}
\usepackage{url}
\usepackage{verbatim}
\usepackage{cite}
\usepackage{tabularx}
\usepackage{booktabs}
\usepackage{multirow}

\usepackage{calrsfs}
\DeclareMathAlphabet{\pazocal}{OMS}{zplm}{m}{n}

\newcolumntype{Y}{>{\raggedleft\arraybackslash}X}

\usepackage{amssymb}
\usepackage{amsthm}
\usepackage{algpseudocode}

\usepackage{accents}
\newcommand{\ubar}[1]{\underaccent{\bar}{#1}}

\usepackage{xcolor}

\usepackage{bm}
\usepackage[numbers,sort&compress,square]{natbib}

\usepackage{balance}

\usepackage[colorlinks,urlcolor=blue,linkcolor=blue,citecolor=blue]{hyperref}

\usepackage{color,array}

\usepackage{graphicx}

\begin{document}

\title{Convolutional Neural Network Compression via Dynamic Parameter Rank Pruning} 

\author{
        Manish~Sharma,~\IEEEmembership{Student Member,~IEEE,}
        Jamison~Heard,~\IEEEmembership{Member,~IEEE,}
        Eli~Saber,~\IEEEmembership{Senior Member,~IEEE,}
        Panos~P.~Markopoulos,~\IEEEmembership{Senior Member,~IEEE}

\thanks{
        This research was supported by an academic grant from the National Geospatial-Intelligence Agency Award No. HM0476-19-1-2014, Project Title: Target Detection/Tracking and Activity Recognition from Multimodal Data. Any opinions, findings and conclusions or recommendations expressed in this material are those of the author(s) and do not necessarily reflect the views of NGA, DoD, or the US government. Approved for public release, NGA-U-2023-01817.

        This research was also supported in part by the Air Force Office of Scientific Research (AFOSR) under award FA9550-20-1-0039.
        
        
        M. Sharma is with the Chester F. Carlson Center for Imaging Science, Rochester Institute of Technology, Rochester, NY 14623 USA (e-mail: ms8515@rit.edu). 	
        
        J. Heard is with the Department of Electrical and Microelectronic Engineering, Rochester Institute of Technology, Rochester, NY 14623 USA (e-mail: jrheee@rit.edu).
        	
        E. Saber is with the Department of Electrical and Microelectronic Engineering and with the Chester F. Carlson Center for Imaging Science, Rochester Institute of Technology, Rochester, NY 14623 USA (e-mail: esseee@rit.edu).


        P. P. Markopoulos is with the Department of Electrical and Computer Engineering and Department of Computer Science, The University of Texas at San Antonio, San Antonio, TX 78204 USA (e-mail: panagiotis.markopoulos@utsa.edu).
	}
        
        }



\maketitle

\begin{abstract}
While Convolutional Neural Networks (CNNs) excel at learning complex latent-space representations, their over-parameterization can lead to overfitting and reduced performance, particularly with limited data. This, alongside their high computational and memory demands, limits the applicability of CNNs for edge deployment. Low-rank matrix approximation has emerged as a promising approach to reduce CNN parameters, but its application presents challenges including rank selection and performance loss. To address these issues, we propose an efficient training method for CNN compression via dynamic parameter rank pruning. Our approach integrates efficient matrix factorization and novel regularization techniques, forming a robust framework for dynamic rank reduction and model compression. We use Singular Value Decomposition (SVD) to model low-rank convolutional filters and dense weight matrices and we achieve model compression by training the SVD factors with back-propagation in an end-to-end way. We evaluate our method on an array of modern CNNs, including ResNet-18, ResNet-20, and ResNet-32, and datasets like CIFAR-10, CIFAR-100, and ImageNet (2012), showcasing its applicability in computer vision. Our experiments show that the proposed method can yield substantial storage savings while maintaining or even enhancing classification performance.
\end{abstract}

\begin{IEEEImpStatement}
The rapid proliferation of edge computing and Internet of Things devices demands lightweight yet efficient machine learning models. Current Convolutional Neural Networks (CNNs), while powerful, are often too resource-intensive for these applications. Our proposed Dynamic Parameter Rank Pruning method automates the compression of CNNs during training without compromising performance. Specifically, our approach dynamically adapts the rank of layers during training based on data and task complexity, thereby eliminating the need for meticulous rank selection and manual adjustments either before or after training. This research is crucial for enabling smarter, more resource-efficient applications in diverse fields such as healthcare diagnostics, autonomous driving, and remote sensing. Our approach democratizes access to advanced networks, making them feasible for deployment in resource-constrained environments. Both practitioners and researchers will find this advancement useful in accelerating the widespread adoption of machine learning solutions in real-world scenarios.
\end{IEEEImpStatement}


\section{Introduction}
\label{sec-intro}
\IEEEPARstart{T}{he} versatility of deep Convolutional Neural Networks (CNNs) is well-documented, finding applications in various areas, such as computer vision \cite{simonyan2014very, he2016deep, krizhevsky2017imagenet}, remote sensing \cite{dhanaraj2020vehicle, sharma2020yolors, sharma2021yolors,singh2023multimodal}, medical diagnosis \cite{litjens2017survey}, and autonomous driving \cite{wen2022deep}, among others. CNNs are favored due to their ability to automatically extract features, promote sparsity and weight sharing, and for their end-to-end trainability. As CNNs are increasingly utilized to tackle complex problems, their underlying models have become more sophisticated, employing a large number of trainable parameters in the form of convolutional filters and fully-connected weight matrices \cite{he2016deep}. Although these large-scale models are viable in computer vision applications with abundant training data and resources, they pose challenges in environments with limited training examples or computational resources such as remote sensing and edge computing \cite{denil2013predicting}.

Several model compression techniques have been proposed to address this issue, including knowledge distillation \cite{hinton2015distilling, zagoruyko2016paying, yim2017gift, ahn2019variational, park2019relational}, quantization \cite{gong2014compressing, han2015deep, courbariaux2016binarized, lin2016fixed, zhouincremental}, pruning \cite{han2015deep, molchanov2016pruning, lipruning}, and special convolution operations \cite{iandola2016squeezenet, chollet2017xception, howard2017mobilenets}. The majority of these methods target pretrained models, and often do not prioritize compression during training. This can lead to a degradation in model performance after compression and typically necessitates retraining. In contrast, low-rank factorization methods offer a promising model compression approach \cite{sharma2021yolors, sharma2021convolutional, hyder2022incremental} as they approximate weight matrices/convolutional filters with low-rank matrix/tensor factors, yielding efficient model compression \cite{denil2013predicting, denton2014exploiting, cp, tuc_conv, ult_tensor}. Nonetheless, the successful deployment such a low-rank factorized model necessitates meticulous rank selection that is tailored to the baseline model architecture and the complexity of the data or task at hand. Certainly, it is a hard or infeasible task to successfully select rank before training.  

Low-rank matrix factorization approaches can, in general, be divided into three categories: (i) post-training low-rank factorization followed by pruning and fine-tuning \cite{denton2014exploiting, cp, tuc_conv, ult_tensor}; (ii) low-rank factorization prior to training with a fixed architecture \cite{sainath2013low, sharma2021yolors}; and (iii) models factorized prior to training with an adaptable architecture approach \cite{chung2019parameter, xu2021trp, chu2021low, yin2022batude}. The third category has recently attracted interest due to its ability to leverage redundancies in trainable parameters during training, thus saving computational resources. However, methods in this category often have limited applicability in terms of the type of layer  they act on, require post-training interventions and fine-tuning or retraining, and can lead to improper convergence and performance deterioration.

To address the above in this work, we introduce \emph{Dynamic Parameter Rank Pruning (DPRP)}, a novel training method that compresses a CNN in an automated way, while training, via dynamic adaptation of the rank of its parameters. Our proposed method employs Singular Value Decomposition (SVD) in conjunction with novel parameter matrix reshaping to model the convolutional filters and dense weight matrices. This integration is facilitated through our proposed regularizations, which impose explicit SVD conditions during training. These regularizations promote orthogonality, the sorting of singular values in decreasing order of importance, and sparsity in the minor singular values, which facilitates rank reduction. Utilizing back-propagation, instead of the weigh matrices or convolutional filters, we directly train their SVD factors, thereby integrating compression directly into the training pipeline. That is, in contrast to the state of the art, our approach dynamically determines the rank of the factorized matrices \emph{during training}, enabling it to adapt to specific task requirements and achieve higher compression rates while maintaining or even enhancing model performance.

The remainder of this paper is organized as follows. Section \ref{sec-related} offers a comprehensive literature review on network compression. Our proposed method is presented in Section \ref{sec-proposed}, followed by extensive experimental studies in Section \ref{sec-experiments}. Subsequent Sections \ref{sec-discussion} and \ref{sec-conclusions} present discussions and concluding remarks, respectively.

\section{Related Work}
\label{sec-related}
In the literature, numerous techniques have been proposed to address the model compression. A prominent approach is knowledge distillation, where a large, accurate model (the teacher) guides a smaller model (the student) by an appropriate transfer of knowledge \cite{hinton2015distilling}. Although this technique improves the efficiency of the student model by leveraging the rich representations learned by the teacher model, most current methods focus on distilling knowledge after the teacher model has been trained \cite{zagoruyko2016paying, yim2017gift, ahn2019variational, park2019relational}, potentially missing opportunities for compression during the training process itself.

Quantization, another model compression technique, reduces the precision of network parameters and activations to decrease memory footprint and accelerate computations \cite{gong2014compressing, han2015deep, courbariaux2016binarized}. However, these techniques struggle to balance quantization-induced loss while maintaining sufficient model capacity; and most methods focus on post-training quantization \cite{lin2016fixed, zhouincremental}, leaving the potential for exploring in-training quantization that allows for simultaneous compression. 

Pruning techniques have also been employed for model compression by identifying and removing redundant or less important parameters \cite{han2015deep}. However, these techniques usually involve an iterative process of pruning and subsequent fine-tuning, which can be computationally expensive \cite{liu2017learning}. Despite the majority of pruning methods being implemented post-training \cite{molchanov2016pruning, lipruning}, some recent approaches have considered pruning during the training phase, predominantly concentrating on enforcing sparsity or binary weights \cite{guo2016dynamic, huang2018data}. 

There has been interest in specially designed convolutional layers, such as depth-wise separable convolutions, for their potential to reduce model complexity. These layers aim to factorize standard convolutions into separate depth-wise and point-wise convolutions, decreasing the number of parameters and operations. However, current studies primarily focus on replacing standard convolutions in predefined architectures \cite{iandola2016squeezenet, chollet2017xception, howard2017mobilenets}, leaving unexplored research space for adaptive and dynamic integration of such layers during training.

Low-rank factorization approaches play a vital role in model compression by reducing the architecture and size of the factorized model \cite{kossaifi2017tensor, tran2018improving, kossaifi2019t, panagakis2024tensor}. Depending on the operational characteristics of low-rank matrix factorization, these methods can, in general, be divided into three categories. The first category involves post-training low-rank factorization followed by pruning and fine-tuning \cite{denton2014exploiting, cp, tuc_conv, ult_tensor, gusak2019automated}. Similar to other model compression techniques, these methods do not prioritize model compression during training, leading to a performance decline after pruning. Extensive retraining is required to restore model performance. The second category is defined by low-rank factorization before training with a fixed architecture \cite{sainath2013low, sharma2021yolors}. In this approach, the low-rank factors are trained during the training phase, making these methods more resilient to performance degradation after pruning, and thus requiring less retraining for fine-tuning. However, determining the appropriate ranks for factorization in both these methods requires considerable effort/time and multiple iterations. Moreover, enforcing a uniform compression rate across all network layers is inefficient, as different layers exhibit varying degrees of redundancies and susceptibility to compression. This uniform low-rank strategy often leads to deteriorated performance. There are methods that emerge as a mixture of the above two approaches, \cite{yang2020learning}, utilizing training with full-rank decomposition while maintaining SVD conditions in the process followed by post-training singular values pruning and fine-tuning to recover the degraded performance.

Recently, attention has shifted towards the third category of low-rank factorization, which involve factorizing models before training with an adaptable architecture approach \cite{chung2019parameter, xu2021trp, chu2021low, yin2022batude}. In this approach, models are generally factorized initially with full rank. During the training process, the factors are gradually transformed into low-rank structures. These methods exploit redundancies in trainable parameters during training, eliminating the need for post-training fine-tuning thereby saving effort, time, and computational resources. To this effect, one study \cite{chung2019parameter} applied this approach to speech recognition, wherein only the fully-connected layers were factorized with actual model compression conducted post-training.

Within image classification, \cite{frankle2018lottery} suggested the use of rank-adaptive evolution on a low-rank manifold for training and compression of networks. This approach, interestingly, avoids the need for full weight representation but it was limited to matrix-valued layers only. In another attempt, Tucker-2 decomposition was used to factorize convolutional layers with regularization gates and funnel function to determine suitable ranks \cite{chu2021low}. However, model compression was implemented post-training followed by a fine-tuning stage that incorporated the evaluation of computational costs relative to the original baseline model, layer swapping, and training of the resultant network from scratch.
Another study proposed a budget-aware Tucker-2 compression approach taking model size constraints into account \cite{yin2022batude}. Imposing stringent constraints on model capacity during the training phase showed an improper convergence in the rank and accordingly in the number of trainable parameters over the course of training across different layers.
With a new training strategy that alternates between low-rank approximation and standard training after a set number of optimization iterations, Tensor Rank Pruning (TRP) \cite{xu2021trp} exploits both space-wise \cite{jaderberg2014speeding} and channel-wise \cite{zhang2015accelerating} correlations to decompose convolutional filters. Unlike the approach of training from scratch, this method is employed during training.

However, these investigations indicate a research gap for a truly dynamic model compression method without the need of post-training retraining and fine-tuning. An ideal approach would incorporate an adaptive rank determination mechanism that trains from scratch and is contingent on both baseline model complexity and the complexity of the data or task at hand. \emph{Exactly this is the gap that we  fill in this work}.

Furthermore, based on the type of factorization, low-rank factorization methods, in general, can be categorized into matrix and tensor methods \cite{chung2019parameter, phan2020stable, chu2021low, yin2022batude}. While some tensor-based low-rank factorization methods provide a wider scope for compression \cite{kim2015compression, astrid2017cp, li2021heuristic, zangrando2023rank}, they often require the determination of multiple ranks per layer in the network, making their appropriate selection a tedious task. Therefore, our proposed approach utilizes the SVD matrix factorization method.

\section{Proposed Method}
\label{sec-proposed}

\begin{figure}[t!]
    \centering
    \includegraphics[width=0.9\linewidth, trim={9.2cm 9.1cm 9.2cm 9.1cm},clip]{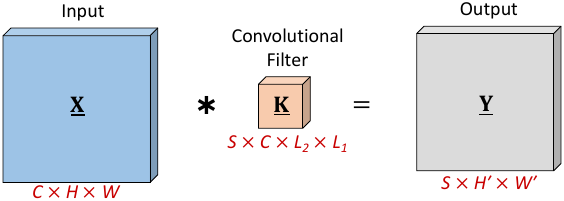}\vspace{-0.2cm}
    \caption{A typical convolutional layer.}
    \label{conv}
\end{figure}

\begin{figure}[t!]
    \centering
    \includegraphics[width=0.56\linewidth, trim={10.8cm 9.06cm 11.0cm 9.08cm},clip]{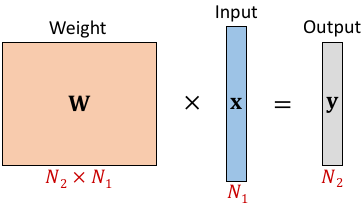}\vspace{-0.2cm}
    \caption{A typical fully-connected layer.}
    \label{fc}
\end{figure}

CNNs primarily consist of convolutional and fully-connected layers. In a convolutional layer, as shown in Fig. \ref{conv}, trainable parameters reside in the convolutional filter. In a fully-connected layer, as shown in Fig. \ref{fc}, trainable parameters are arranged in dense weight matrix. In this work, we demonstrate how SVD matrix factorization, coupled with proposed regularizations, can effectively model these elements of deep CNNs for dynamic compression via parameter rank updates during training. This, in turn, reduces redundancy and enhances performance, even when applied to optimized, efficient, standard and state-of-the-art deep CNNs. 

\subsection{Notation and SVD Preliminaries}
Throughout this paper, we adhere to the following notation: scalar variables are represented by lowercase letters (e.g., $x$), vectors are indicated by boldface lowercase letters (e.g., $\mathbf{x}$), matrices are denoted by boldface uppercase letters (e.g., $\mathbf{X}$), and tensors are signified by underscored boldface uppercase letters (e.g., $\mathbf{\ubar{X}}$). The identity matrix is symbolized by $\mathbf{I}$, and real numbers are signified by $\mathbb{R}$. To represent the entries of a vector, matrix, or tensor, we use the notation $[\cdot]_i$, where $i$ denotes a set of indexes. $\mathbf{X}^T$ denotes the transpose of $\mathbf{X}$.

Compact SVD, also referred to as SVD in this paper, is a powerful mathematical technique extensively utilized across various domains, including dimensionality reduction, data compression, and collaborative filtering \cite{golub2013matrix, chung2019parameter}. It decomposes a matrix into: the left singular vectors $\mathbf{U}$, the singular values ($\boldsymbol{\sigma}$) in diagonal matrix $\mathbf{\Sigma}$, and the transposed right singular vectors $\mathbf{V}^T$. In mathematical terms, given $\mathbf{A} \in \mathbb{R}^{m \times n}$ of rank $r$, the SVD factorization is expressed as $\mathbf{A} = \mathbf{U} \mathbf{\Sigma} \mathbf{V}^T$, where $\mathbf{U} \in \mathbb{R}^{m \times r}$, $\mathbf{\Sigma} \in \mathbb{R}^{r \times r}$, and $\mathbf{V}^T \in \mathbb{R}^{r \times n}$. SVD features several crucial properties such as orthogonality, whereby $\mathbf{U}$ and $\mathbf{V}$ are orthogonal matrices, i.e., $\mathbf{U}^T\mathbf{U}=\mathbf{I}$ and $\mathbf{V}^T\mathbf{V}=\mathbf{I}$, meaning their columns form an orthonormal basis. Moreover, $\boldsymbol{\sigma}$ are non-negative and are arranged in descending order, thereby enabling the identification of the most significant components in the matrix. The rank of the matrix can be discerned by examining the number of non-zero singular values, offering insights into the inherent structure and dimensionality of the original matrix.

\subsection{Factorized Convolutional and Fully-Connected Layer}

\subsubsection{Convolutional Layer}
Consider convolutional filter $\mathbf{\ubar{K}} \in \mathbb{R}^{S \times C \times L_2 \times L_1}$. It is a $4$-way tensor comprising $S$ $3$-way kernels of pixel width $L_1$, pixel height $L_2$, and channel depth $C$. Each kernel convolves with an input image $\mathbf{\ubar{X}} \in \mathbb{R}^{C \times H \times W}$, which is again a $3$-way tensor of pixel width $W$, pixel height $H$, and channel depth $C$. The convolution is performed with padding parameters $(p_1, s_1)$ and $(p_2, s_1)$, controlling padding and stride along the width and height of $\mathbf{\ubar{X}}$, respectively. The result of the convolution is a $3$-way output tensor $\mathbf{\ubar{Y}} = \mathbf{\ubar{X}} \ast \mathbf{\ubar{K}} \in \mathbb{R}^{S \times H' \times W'}$, where  $W'=(W-L_1+2p_1)/s_1 + 1$ and $H'=(H-L_2+2p_2)/s_2+1$, as shown in Fig. \ref{conv}. In the case of symmetric convolution, which is typically the case, $L_1=L_2=L$, $p_1=p_2=p$, and $s_1=s_2=s$. The number of trainable parameters contained in a standard convolutional filter is $P_{\text{c}} = SCL_1L_2$.

To factorize a convolutional layer, we first consider reshaping of tensor $\mathbf{\ubar{K}}$ into matrix $\mathbf{M} \in \mathbb{R}^{SC \times L_1L_2}$ so that 
\begin{equation}
[\mathbf{\ubar{K}}]_{s,c,l_2,l_1}=[\mathbf{M}]_{i,j},
\end{equation}
where $i=(s-1)C+c$ and $j=(l_2-1)L_1+l_1$ with $s=1,2,\ldots,S$, $c=1,2,\ldots,C$, $l_2=1,2,\ldots,L_2$, and $l_1=1,2,\ldots,L_1$. Next, we consider that $\mathbf M$ is of rank $r \leq \min\{SC, L_1L_2\}$, attaining SVD $\mathbf M = \mathbf{U} \mathbf{\Sigma} \mathbf{V}^T$, so that
$
[\mathbf{M}]_{i,j} = \sum_{g=1}^{r} [\mathbf{U}]_{i,g}  [\mathbf{\Sigma}]_{g,g}  [\mathbf{V}]_{j,g},
$
where $i=1,2,\ldots,SC$ and $j=1,2,\ldots,L_1L_2$. Thus, effectively, through the low-rank structure of $\mathbf M$, convolutional filter $\mathbf K$ is factorized as
\begin{equation}
[\mathbf{\ubar{K}}]_{s,c,l_2,l_1}=\sum_{g=1}^{r} [\mathbf{U}]_{(s-1)C+c,g}  [\mathbf{\Sigma}]_{g,g}  [\mathbf{V}]_{(l_2-1)L_1+l_1,g}
\end{equation}
for every $s=1,2,\ldots,S$, $c=1,2,\ldots,C$, $l_2=1,2,\ldots,L_2$, and $l_1=1,2,\ldots,L_1$. The particular reshaping/matricization of $\mathbf{\ubar{K}}$ to $\mathbf{M}$ was selected in order to reduce the number of trainable parameters and computational overhead. Instead of training the entries of $\mathbf{\ubar{K}}$, we train the entries of its factors in $\mathbf{U} \in \mathbb R^{SC \times r}$, $\mathbf V \in \mathbb R^{L_1L_2 \times r}$, and $\mathbf{\Sigma} \in \mathbb R^{r \times r}$. Thus, the number of trainable parameters in a factorized convolutional layer is given by $P_{\text{fc}} = r(SC + L_1L_2 +1)$. Accordingly, the  proposed factorization constitutes parameter compression when $P_{\text{fc}} \leq P_{\text{c}}$ or, equivalently,
\begin{equation}
    r \leq \frac{SCL_1L_2 }{SC + L_1L_2 + 1}.
\end{equation}
The corresponding compression rate, as a function of $r$, is
\begin{equation}
    R_{\text{fc}}(r) = 1-\frac{P_{\text{fc}}}{P_{\text{c}}} = 1- \frac{r(SC + L_1L_2 + 1)}{SCL_1L_2}.
\end{equation}

\subsubsection{Fully-Connected Layer}
In the case of a fully-connected layer, a dense weight matrix $\mathbf{W} \in \mathbb{R}^{D_2 \times D_1}$ is multiplied with input $\mathbf{x} \in \mathbb{R}^{D_1}$ resulting in the output $\mathbf{y} = \mathbf{W}\mathbf{x} \in \mathbb{R}^{D_2}$. The number of trainable parameters in a standard fully-connected layer is given by $P_{\text{f}} = D_1D_2$.
For a factorized fully-connected layer, $\mathbf{W}$ is considered to be of low rank $r \leq \min \{D_1, D_2 \}$, admitting SVD $\mathbf W=\mathbf U \mathbf{\Sigma} \mathbf V^T$, so that 
\begin{equation}
[\mathbf{W}]_{d_2,d_1} = \sum_{g=1}^{r} [\mathbf{U}]_{d_2,g}  [\mathbf{\Sigma}]_{g,g}  [\mathbf{V}]_{d_1, g},
\end{equation}
for $d_2=1,2,\ldots,D_2$ and $d_1=1,2,\ldots,D_1$. That is, instead of learning $\mathbf{W}$, the proposed method learns the SVD factors in $\mathbf{U} \in \mathbb R^{D_2 \times r}$, $\mathbf V \in \mathbb R^{D_1 \times r}$, and $\mathbf{\Sigma} \in \mathbb R^{r \times r}$. Accordingly, the number of trainable parameters in a factorized fully-connected layer is given by $P_{\text{ff}} = r(D_1 + D_2 + 1)$. For the factorization to accomplish compression we need $P_{\text{ff}} \leq P_{\text{f}}$ or, equivalently,
\begin{equation}
    r \leq \frac{D_1 D_2 }{D_1 + D_2 + 1}.
\end{equation}
The attained  compression rate, as a function of $r$, is  
\begin{equation}
    R_{\text{ff}}(r) = 1-\frac{P_{\text{ff}}}{P_{\text{f}}} = 1- \frac{r(D_1 + D_2 + 1)}{D_1 D_2}.
\end{equation}
 
\begin{figure}[t!]
    \centering
    \includegraphics[width=1\linewidth, trim={8.6cm 6.3cm 8.2cm 6.3cm},clip]{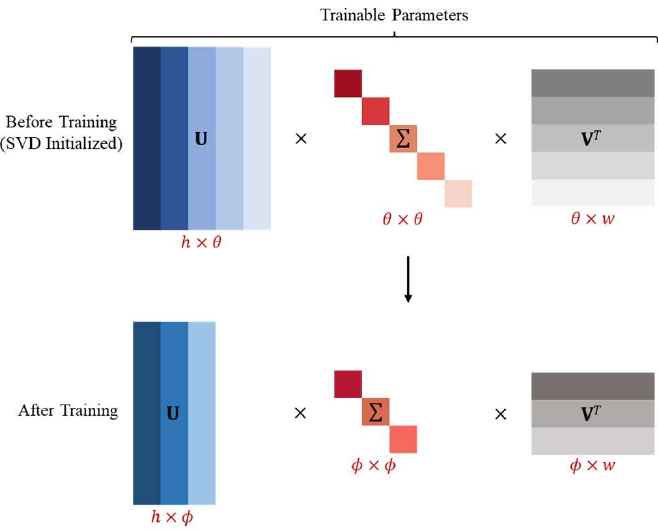}
    \caption{The variation in factor sizes, represented by the initial rank $\theta$ and the final rank $\phi$. }
    \label{training}
\end{figure}

\subsection{Factor Initialization and Training}

Below we present the proposed training of the parameter factors of a layer, whether convolutional or fully-connected. For ease in notation, we denote $(h=SC, w=L_1L_2)$ if the layer is convolutional or $(h=D_2, w=D_1)$. We begin the model training with SVD-factorized convolutional and fully-connected layers of full-rank $\theta = \min \{ h, w \}$, as illustrated in the top-half of Fig. \ref{training}. 

The SVD structure (orthonormality of singular-vectors and sortment singular values) and preferred low rank are determined implicitly throughout training via intelligently designed loss functions.  Accordingly, the total loss function considered for training is 
\begin{equation}
    L_{\text{total}} = L_{\text{app}} + \lambda_\text{str} L_{\text{str}} + \lambda_\text{comp} L_{\text{comp}}, \label{totalloss}
\end{equation}
where $\lambda_\text{str}$ and $\lambda_\text{comp}$ are loss-weighing hyperparameters.
$L_{\text{app}}$ is the loss function pertinent to the application at hand (e.g., image classification, object detection, segmentation). $L_{\text{str}}$ is the loss responsible for maintaining the SVD structure of the parameter factorizations across the layers. Finally, $L_{\text{comp}}$ is the function responsible for promoting parameter rank reduction and, thus, model compression. Next, we present the three proposed losses in detail. 

\subsubsection{Application Loss}
This loss is pertinent to the application at hand and can vary across CNN deployments. For the sake of numerical experimentation, in this paper we consider an entropy-based classification loss 
\begin{equation}
L_{\text{app}} = -\frac{1}{n_c} \sum_{n=1}^{n_c} y_n \ln{(\hat{y}_n)} \label{eq:class_loss},
\end{equation}
where $n_c$ denotes the number of classes in the classification task, $y_n$ indicates the ground-truth, and $\hat{y}_n$ represents the prediction. It is worth noting that this loss term strives to improve classification performance on the training data, regardless of factor structure and compression, which will have to be regulated by the two loss terms presented below. 

\subsubsection{Structure Loss}
Next, we create a loss term that promotes SVD structure and, thus, facilitates adaptive rank and parameter compression. We recognize that there are two main components in the SVD structure: (i) orthonormality of the singular vectors and (ii) sortment of the singular values. Accordingly, we analyze $L_{\text{str}}$ in two corresponding sub-terms: $L_{\text{str}} = \mu_{\text{orth}} L_{\text{orth}} + \mu_{\text{sort}} L_{\text{sort}}$, where $\mu_{\text{orth}}$ and $\mu_{\text{sort}}$ are hyper-parameter weights.
Denoting by $\{ \mathbf U_{l}, \mathbf{\Sigma}_l, \mathbf V_l\}$ and $r_l$ the SVD-factors and SVD-rank for layer $l$, respectively, we define 
\begin{equation}
\hspace{-0.2cm} L_{\text{orth}} = \frac{1}{L} \sum_{l=1}^L \frac{1}{r_l^2}  \left( \|\mathbf{U}_l^T\mathbf{U}_l-\mathbf{I}\|_F + \|\mathbf{V}_l^T\mathbf{V}_l-\mathbf{I}\|_F  \right) \label{eq:ortho_loss}.
\end{equation}
This loss term promotes orthogonality to the left- and right-hand singular matrices, across all layers, with an emphasis normalized by each layer's rank.

Next, we design a loss term that promotes sortment of the singular values in $\{\mathbf \Sigma_l\}_{l=1}^L$ so that dynamic truncation could result to  optimal low-rank approximation, in accordance with the SVD principles. Specifically, $L_{\text{sort}}$ strives to accomplish $[\boldsymbol{\sigma}_l]_j \geq [\boldsymbol{\sigma}_l]_{j+1} \geq 0 ~ \forall \; j \in [1,r_l)$ and $\forall \; l \in \{1,2,\ldots,L \}$. Let the set $I_l$ contain the indices of all singular values of layer $l$ that are 
out of desired order;
that is, $I_l = \{j \in \{ 2,\ldots, r_l\}:~ [\boldsymbol \sigma_l]_{j}>[\boldsymbol \sigma_l]_{j-1}\} $. Accordingly, define the cardinality (number of entries) of $I_l$ as $\gamma_l=|I_l|$. Also, let $\eta_l$ denote the number of negative entries in $\mathbf \sigma_l$. Moreover, define function $\chi: \mathbb N \rightarrow \mathbb R_0^+$ such that, for every $a \in \mathbb N$, $\chi(a)=1/a$, if $a>0$, and $\chi(a)=0$, if $a=0$. Then, we define the sorting loss term as 
\begin{equation}
\begin{split}
L_{\text{sort}} = \frac{1}{L} \sum_{l=1}^L & \chi(\gamma_l) \sum_{j=1}^{r_l-1} \max \{ 0, [\boldsymbol{\sigma}_l]_{j+1}-[\boldsymbol{\sigma}_l]_j \} \; \\ + &  \chi(\eta_l) \sum_{j=1}^{r_l-1}  \max \{ 0, -[\boldsymbol{\sigma}_l]_j \} \label{eq:sort_loss}.
\end{split}
\end{equation}
The scaling terms $\chi(\gamma_l)$ and $\chi(\eta_l)$ are used so as to prevent layers with large $\gamma_l$ and $\eta_l$, respectively, from dominating the loss. Overall, $L_{\text{sort}}$ promotes that, across $l$, the entries of $\boldsymbol{\sigma}_l$ are non-negative and arranged in descending order. 
 
 \subsubsection{Compression Loss}
 To facilitate dynamic compression we perform dynamic rank reduction. We denote by $\tau_l$ the reduced rank of layer $l$ as the highest value of  $i$ for which $|[\boldsymbol{\sigma}_l]_{i+1}| > \epsilon |[\boldsymbol{\sigma}_l]_{i}|  \; \forall \; i \in [1, r_l)$, for some pruning threshold $\epsilon \in (0,1)$ (hyper-parameter). Then, we perform rank reduction by removing all singular values $\{[\boldsymbol \sigma_l]_j\}_{j>\tau_l}$ (see Section \ref{sec-compression} below). To make sure that this pruning comes with minimum approximation loss, we promote sparsity in  $\{[\boldsymbol \sigma_l]_j\}_{j>\tau_l}$ by means of the compression loss term:
\begin{equation}
L_\text{comp} = \frac{1}{L} \sum_{l=1}^L \frac{1}{(r_l-\tau_l)||\boldsymbol{\sigma}_l||_2} \sum_{i=\tau_l}^{r_l} |[\boldsymbol{\sigma}_l]_i| \label{eq:diag_loss}.
\end{equation}
 In $L_\text{comp}$, we  divide by $||\boldsymbol{\sigma}_l||_2$ in order to prevent layers with relatively larger minimal singular values across layers to dominate the regularization term. Also, we divide by $r_l-\tau_l$ in order to avoid domination by layers with a relatively large number of singular values to be reduced. This arrangement promotes pruning of the minimal singular values, facilitating model compression through dynamic rank reduction in training. 

\subsection{Model Compression}
\label{sec-compression}
While training, for a given layer $l$, we dynamically reduce the value of $r_l$ to $\tau_l$ by removing $[\boldsymbol{\sigma}_l]_{i=\tau_l+1}^{r_l}$. Accordingly, the corresponding trainable parameters are removed from $\mathbf{U}_l$ and $\mathbf{V}^T_l$. Since the removed singular values have been reduced throughout training, their influence on the final convolution filter is minimal. Thus, their removal does not significantly affect performance. If $\tau_l = r_l$, no trainable parameters are removed and the network continues training with the same number of trainable parameters as before.
At the end of training $\tau_l=\phi_l$, resulting in a compact model, as illustrated in the bottom-half of Fig. \ref{training}.

\section{Experimentation}
\label{sec-experiments}
In this section, we detail the experimental datasets, baseline models, evaluation metrics, experimental configurations, and results obtained for the proposed method in comparison to baselines and other comparative approaches for the image classification applications.

\subsection{Datasets, Baseline Models, and Evaluation Metrics}
Our image classification experiments utilize three common computer vision datasets: CIFAR-10, CIFAR-100, and ImageNet (2012) \cite{russakovsky2015imagenet}, consisting of $10$, $100$, and $1000$ classes, respectively. CIFAR-10 and CIFAR-100 datasets both contain $50$K training and $10$K testing images of $32\times32$ resolution. For both datasets, samples are uniformly distributed across classes in the train and test sets. The ImageNet dataset, on the other hand, contains approximately $1.2$M training images, $50$K validation images, and $150$K testing images with an average resolution of $469 \times 387$. Due to the absence of ground-truth for the test set, the validation set is utilized for testing. Standard transformations and augmentations techniques are employed to increase data variation in an online manner and provide a larger diverse dataset while training \cite{krizhevsky2012imagenet, he2016deep}.

Baseline models for the CIFAR-10 and CIFAR-100 datasets utilize ResNet-20 and ResNet-32 networks, respectively. On the contrary, the ImageNet dataset employs ResNet-18 network as its baseline models \cite{he2016deep}. ResNet-20 and ResNet-32 are generally considered smaller networks suitable for CIFAR-10 and CIFAR-100 datasets.

We employ Top-1 and Top-5 accuracies as our primary evaluation metrics for classification performance. Top-1 accuracy is the percentage of times the model correctly predicts the highest ranked class, whereas Top-5 accuracy is the percentage of times the top 5 predictions of the model include the correct class. In addition, MMAC (Mega Multiply-Accumulate operations per second) and GMAC (Giga Multiply-Accumulate operations per second) are used to gauge a model computational complexity, with smaller MMAC/GMAC values denoting faster models. 

For comparative methods, in case of code unavailability, results are directly sourced from the corresponding publications. Since we train our baseline model from scratch similar to methods \cite{ning2020dsa, xu2021trp, chu2021low}, so, our baseline accuracy differs from the comparative method that utilize Torchvision pre-trained weights \cite{li2021heuristic, yin2022batude} for baseline accuracy. Thus, for a fair comparison, if the baseline accuracy in the source, $A'_{\text{source}}$, differs from our calculated baseline accuracy, $A'_{\text{ours}}$, resulting from use of pre-trained weights or the randomness in model initialization and other non-deterministic uncertainties, we adopt a scaling method as done in \cite{chu2021low} to adjust the comparative accuracy $A_{\text{source}}$, resulting in the scaled accuracy 
\begin{equation}
A_{\text{scaled}} = \frac{A'_{\text{ours}}}{A'_{\text{source}}}  A_{\text{source}}.
\end{equation}

\subsection{Experimental Configuration}
We undergo training for ResNet-20, ResNet-32 and ResNet-18 until convergence is observed in the train-test losses. This was accomplished with over 300 epochs for ResNet-20 and ResNet-18, and 150 epochs for ResNet-18. Each network is trained with a batch size of $256$ images. The training follows the method detailed in \cite{he2016deep} which utilizes the stochastic gradient descent optimizer with a momentum of $0.9$, a weight decay of $1e-4$, and an initial learning rate of $0.1$.
We incorporate a commonly used reduce-on-plateau strategy applied to the classification loss. This strategy involves reducing the learning rate by a factor of $0.1$ when the loss does not decrease within a patience interval of $10$ epochs, allowing the training to continue with the reduced learning rate.
In the case of factorized models, we empirically set ${\lambda}_{\text{str}}=1$, ${\mu}_{\text{orth}}=1000$, and ${\mu}_{\text{sort}}=1$, respectively. Similarly, the values of ${\lambda}_{\text{comp}}$ and $\epsilon$ are also empirically determined, with the actual values contingent upon the specific dataset and baseline model in use, as delineated in Table \ref{tab:lambda_eps}.

\begin{table}[!t]
\caption{Values of hyper-parameters $\lambda$ and $\epsilon$ for best performance using different datasets and baseline models. \label{tab:lambda_eps}}
\centering
\fontsize{6}{8.5}\selectfont
\begin{tabularx}{\linewidth}{XXYY}
\toprule
\textbf{Dataset} & \textbf{Model} & ${\lambda}_{\text{comp}}$ & $\mathbf{\epsilon}$\\
\hline
CIFAR-10 & ResNet-20 & $0.1$ & $0.1$\\
CIFAR-10 & ResNet-32 & $0.5$ & $0.001$ \\
CIFAR-100 & ResNet-20 & $0.1$ & $0.1$\\
CIFAR-100 & ResNet-32 & $1.0$ & $0.001$\\
ImageNet & ResNet-18 & $0.5$ & $0.001$\\
\bottomrule
\end{tabularx}
\end{table}

\subsection{Results}

\subsubsection{Performance Analysis on CIFAR-10 Dataset}
We compare our proposed method with a baseline and several contemporary methods using the CIFAR-10 dataset on ResNet-20 and ResNet-32 networks. The results are tabulated in Table \ref{tab:cifar10}, focusing on Top-1 classification accuracy and the degree of compression in the number of trainable parameters. Two different $\lambda_\text{comp}$ and $\epsilon$ configurations of the proposed method are presented, namely, proposed 1 and proposed 2. For the ResNet-20 based models, we use $\lambda_\text{comp}=0.5$ and $\epsilon=0.01$ for proposed 1, and proposed 2 uses $\lambda_\text{comp}$ and $\epsilon$ values listed in Table \ref{tab:lambda_eps}. For the ResNet-32 based models, we use $\lambda_\text{comp}=1$ and $\epsilon=0.001$ for proposed 1, and again proposed 2 uses the values from Table \ref{tab:lambda_eps}.

Our observations reveal that both configurations of the proposed method provide the highest Top-1 accuracy for ResNet-20 and ResNet-32 at $90.99\%$ and $92.16\%$, and at $92.25\%$ and $93.03\%$ respectively, while simultaneously reducing the number of trainable parameters by $30.66\%$ and $5.79\%$ for ResNet-20, and by $52.78\%$ and $24.96\%$ for ResNet-32, in comparison to the baseline. This indicates that our proposed method configurations are more parameter-efficient relative to the baseline ResNet-20 and ResNet-32 models, even if there is a slight degradation in performance for the proposed 1 configurations.
One important observation to note is that unlike other comparative methods, these efficiencies are achieved without the necessity for post-training fine-tuning/retraining, which significantly reduces post-training processing time and effort. Even though methods such as Std. Tucker \cite{kim2015compression, yin2022batude}, PSTR-M \cite{li2021heuristic}, and BATUDE \cite{yin2022batude} display higher parameter compression rates, they come at the expense of lower accuracy scores. This implies a trade-off between model efficiency and performance. The specific compression and accuracy values of the proposed methods suggest a more balanced approach in dealing with this trade-off.

\subsubsection{Performance Analysis on CIFAR-100 Dataset}

Next, we extend our experimental results to the CIFAR-100 dataset, as depicted in Table \ref{tab:cifar100}. The CIFAR-100 dataset, in contrast to CIFAR-10, offers fewer images per class, thus presenting a scenario for image classification in a resource-constrained environment.

\begin{table}[!t]
\caption{Comparison of methods on ResNet-20 and ResNet-32 using CIFAR-10, showing Top-1 accuracy and parameter compression. Best results for each evaluation metric are highlighted in bold text. \label{tab:cifar10}}
\centering
\fontsize{6}{8.5}\selectfont
\begin{tabularx}{\linewidth}{Xrrrr}
\toprule
\multirow{2.5}{*}{\textbf{Method}} & \multicolumn{2}{c}{\textbf{ResNet-20}} & \multicolumn{2}{c}{\textbf{ResNet-32}}\\
\cmidrule(lr){2-3} \cmidrule(lr){4-5}
& \textbf{Top-1 (\%)} & \textbf{Compression (\%)} & \textbf{Top-1 (\%)} & \textbf{Compression (\%)}\\
\midrule
Baseline & $90.98$ & $0.00$ & $92.47$ & $0.00$\\
Std. Tucker \cite{kim2015compression, yin2022batude} & $87.15$ & $61.54$ & $87.67$ & $80.39$\\
PSTR-M \cite{li2021heuristic} & $89.04$ & $\mathbf{85.29}$ & $90.57$ & $\mathbf{82.76}$\\
PSTR-S \cite{li2021heuristic} & $90.53$ & $60.00$ & $91.41$ & $62.96$\\
BATUDE \cite{yin2022batude} & $90.75$ & $61.54$ & $92.15$ & $64.29$\\
Proposed 1 & $90.99$ & $30.66$ & $92.25$ & $52.78$\\
Proposed 2 & $\mathbf{92.16}$ & $5.79$ & $\mathbf{93.03}$ & $24.96$\\
\bottomrule
\end{tabularx}
\end{table}

\begin{table}[!t]
\caption{Comparison of methods on ResNet-20 and ResNet-32 using CIFAR-100, showing Top-1 accuracy and parameter compression. Best results for each evaluation metric are highlighted in bold text. \label{tab:cifar100}}
\centering
\fontsize{6}{8.5}\selectfont
\begin{tabularx}{\linewidth}{Xrrrr}
\toprule
\multirow{2.5}{*}{\textbf{Method}} & \multicolumn{2}{c}{\textbf{ResNet-20}} & \multicolumn{2}{c}{\textbf{ResNet-32}}\\
\cmidrule(lr){2-3} \cmidrule(lr){4-5}
& \textbf{Top-1 (\%)} & \textbf{Compression (\%)} & \textbf{Top-1 (\%)} & \textbf{Compression (\%)}\\
\midrule
Baseline & $65.46$ & $0.00$ & $68.12$ & $0.00$\\
Std. Tucker \cite{kim2015compression, yin2022batude} & $57.58$ & $60.00$ & $59.05$ & $60.00$\\
PSTR-M \cite{li2021heuristic} & $63.68$ & $\mathbf{78.72}$ & $66.79$ & $\mathbf{80.77}$\\
PSTR-S \cite{li2021heuristic} & $66.19$ & $56.52$ & $68.07$ & $58.33$\\
BATUDE \cite{yin2022batude} & $66.73$ & $64.29$ & $68.97$ & $61.54$\\
Proposed 1 & $65.66$ & $21.55$ & $68.75$ & $42.11$\\
Proposed 2 & $\mathbf{67.36}$ & $5.66$ & $\mathbf{69.96}$ & $35.70$\\
\bottomrule
\end{tabularx}
\end{table}

Again, two distinct configurations of our proposed method, denoted as proposed 1 and proposed 2, are presented for comparison. For the ResNet-32 models, we use $\lambda_\text{comp}=0.5$ and $\epsilon=0.01$ for proposed 1, whereas proposed 2 employs $\lambda_\text{comp}$ and $\epsilon$ values specified in Table \ref{tab:lambda_eps}. For the ResNet-32-based models, we adopt $\lambda_\text{comp}=1$ and $\epsilon=0.1$ for proposed 1, and again, proposed 2 uses the values from Table \ref{tab:lambda_eps}.

The results demonstrate that both proposed 1 and proposed 2 configurations yield the highest Top-1 accuracy for ResNet-20, at $65.66\%$ and $67.36\%$, respectively, while simultaneously achieving a parameter compression of $21.55\%$ and $5.66\%$, respectively, compared to the baseline. For ResNet-32, the proposed 2 configuration gives the highest Top-1 accuracy at $69.96\%$ while achieving a parameter compression of $35.70\%$ in comparison to the baseline. 
Proposed 1 configuration outperforms the baseline and most of the comparative methods (with the exception of BATUDE \cite{yin2022batude}) in Top-1 accuracy at $68.75\%$ while simultaneously reducing the number of trainable parameters by $42.11\%$.

Although the compression rates of the proposed configurations on CIFAR-100 are lower than some of the comparative methods, such as PSTR-M \cite{li2021heuristic}, its higher accuracy highlights an important trade-off between compression rates and classification performance. A higher compression rate does not always equate to better classification performance. Notably, the proposed 1 configuration achieves an improvement of $1.90\%$ over the baseline on ResNet-20 and $1.84\%$ on ResNet-32, while significantly reducing the number of trainable parameters (by $5.66\%$ and $35.70\%$ respectively compared to the baseline). Similarly to CIFAR-10, these improvements are achieved without the need for post-training fine-tuning or retraining, thus saving significant post-training processing time and effort. These results confirm the effectiveness of the proposed method for image classification tasks, especially in resource-constrained environments.

\begin{figure}[t!]
    \centering
    \includegraphics[width=1\linewidth, trim={2.3cm 4.2cm 2.3cm 4.2cm},clip]{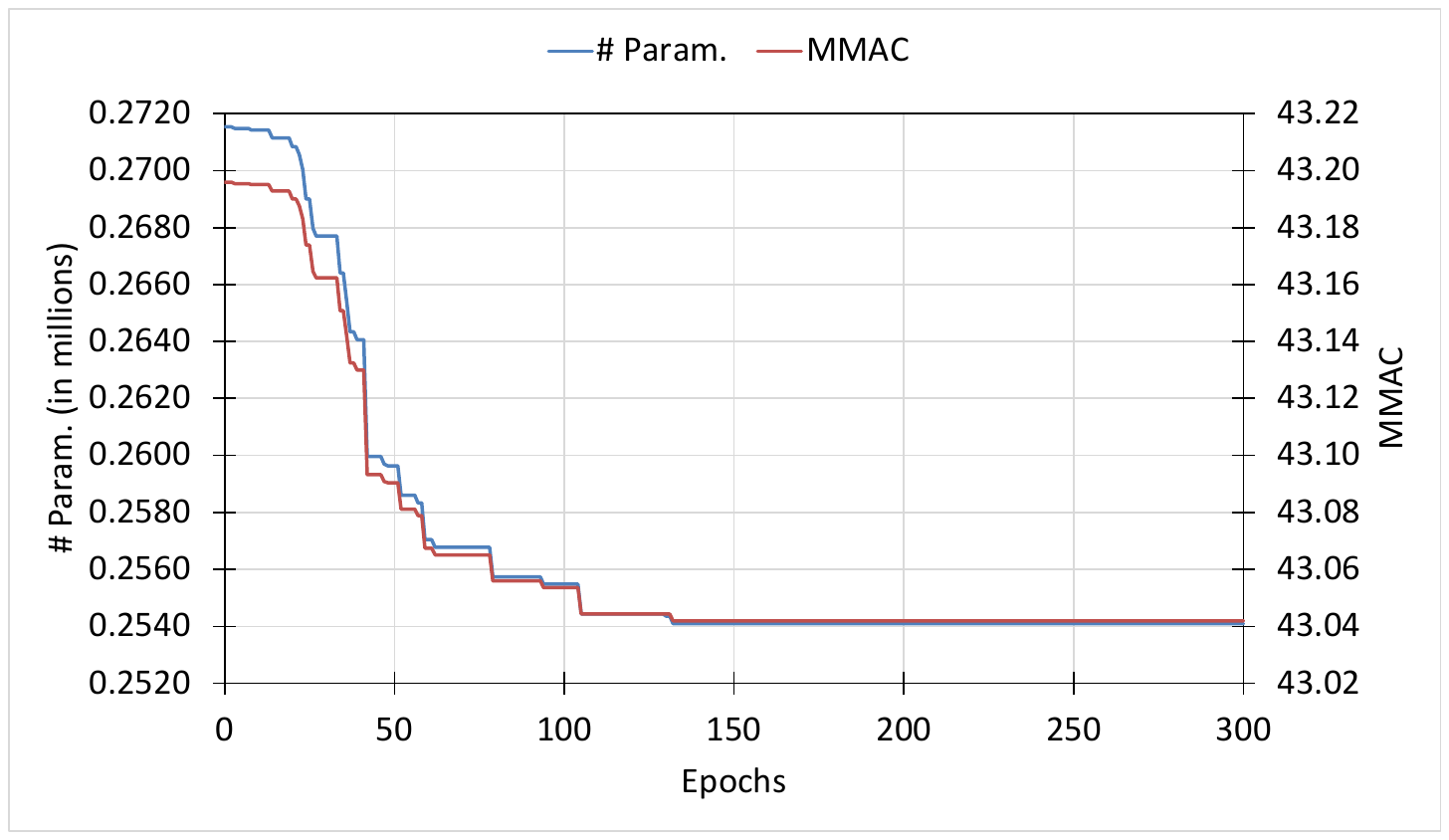}
    \caption{The variation in the number of trainable parameters and FLOPS across epochs for the ResNet-20 proposed model using CIFAR-10 dataset.}
    \label{para_flops}
\end{figure}

\subsubsection{Redundancy Analysis}

Fig. \ref{para_flops} illustrates the variations in the number of trainable parameters and MMAC over the course of training epochs for the ResNet-20 network, utilizing our proposed method on the CIFAR-10 dataset. The plot reveals an initial linear and monotonic decrease in both the number of trainable parameters and MMAC, persisting until approximately the $100$th epoch. Subsequently, a plateau is observed, indicating convergence. This pattern suggests that the original network possessed redundant parameters that were effectively pruned by our proposed method during training. Consequently, a more efficient model was created, improving upon the original architecture, and adapting to the complexity of the data and the task at hand.

\begin{figure}[t!]
    \centering
    \includegraphics[width=1\linewidth, trim={2.4cm 4.3cm 2cm 4.2cm},clip]{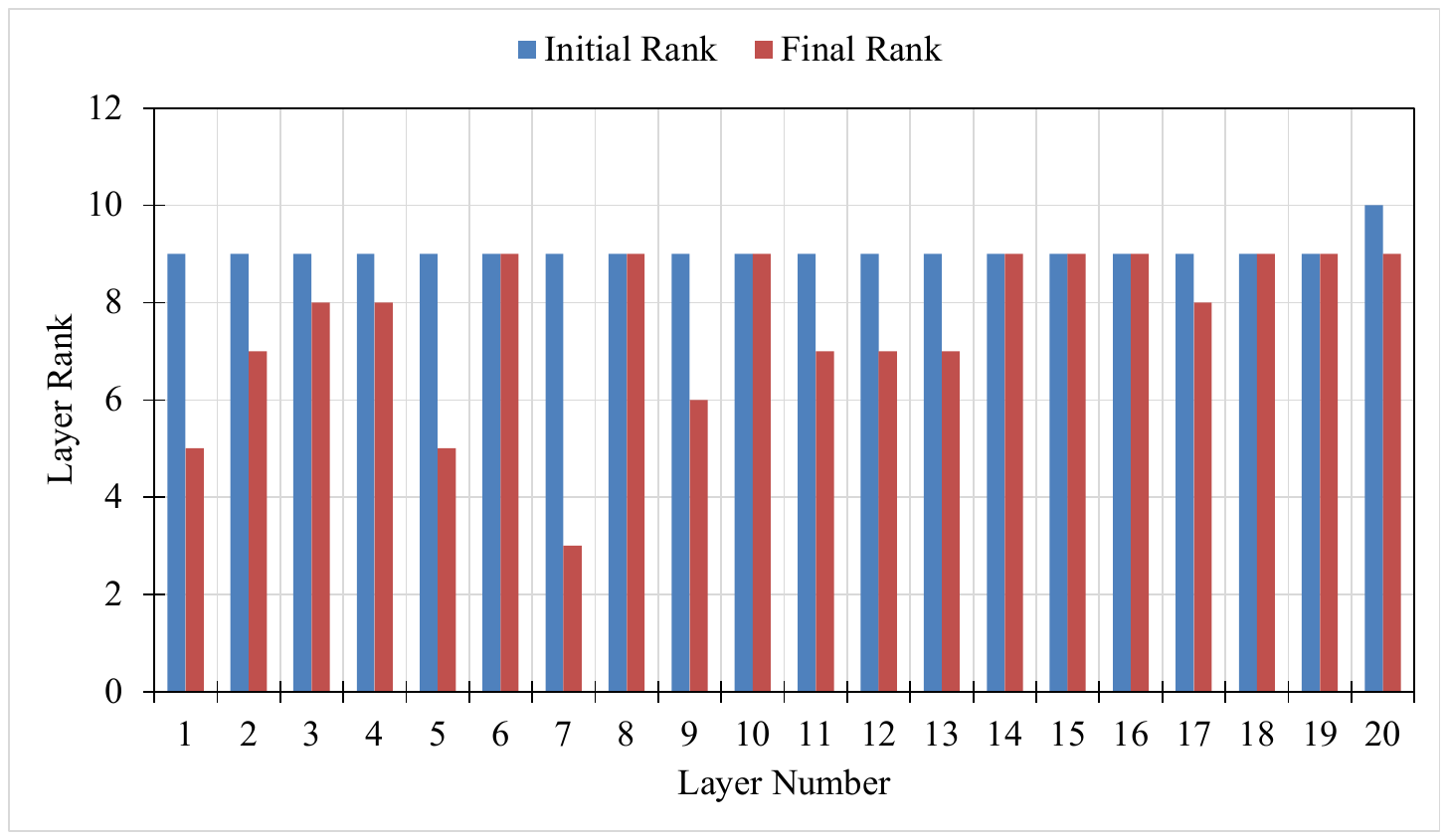}
    \caption{Initial and final rank comparison for the ResNet-20 proposed model using the CIFAR-10 dataset. A smaller rank indicates a more compact layer with relatively fewer trainable parameters.}
    \label{rank_analysis}
\end{figure}

To gauge the degree of rank redundancy across layers in the baseline network, we juxtapose (see Fig. \ref{rank_analysis}) the initial and final ranks of the ResNet-20 network using our proposed method on the CIFAR-10 training dataset. Our observations uncover varying degrees of redundancy, most notably in the early to intermediate layers of the network. These insights pave the way for the design of comparatively leaner networks with fewer trainable parameters per layer.

\subsubsection{Ablation Study with Different Regularizations}
Within this factorization framework, various regularization techniques, such as L1, L2, and funnel \cite{chu2021low}, can be employed for network compression in place of the proposed losses. Similar to \eqref{totalloss}, the general expression for the total loss is of the form
\begin{equation}
    L_\text{total}=L_\text{app} + \lambda_\text{reg} L_\text{reg},
\end{equation}
where $\lambda_\text{reg}$ is the regularization hyper-parameter. For L1 regularization, 
\begin{equation} \label{reg_ablation0}
L_\text{reg} = \frac{1}{L} \sum_{l=1}^L \frac{1}{r_l}||\boldsymbol{\sigma}_l||_1.
\end{equation}
For L2 regularization,
\begin{equation} \label{reg_ablation1}
L_\text{reg} = \frac{1}{L} \sum_{l=1}^L \frac{1}{r_l}||\boldsymbol{\sigma}_l||_2.
\end{equation}
For funnel regularization, 
\begin{equation} \label{reg_ablation}
L_\text{reg} = \frac{1}{L} \sum_{l=1}^L \frac{1}{r_l} \sum_{i=1}^{r_l} \frac{|[\boldsymbol{\sigma}_l]_i|}{|[\boldsymbol{\sigma}_l]_i| + \delta},
\end{equation}
for some low positive value for $\delta$.
Table \ref{tab:reg} summarizes the outcomes of an ablation study that explores the use of different regularization methods as mentioned in (\ref{reg_ablation}) to dynamically facilitate model compression during the training process. The CIFAR-10 dataset on the ResNet-20 baseline network serves as the foundation for this analysis, and each method is evaluated in terms of Top-1 classification accuracy, compression (i.e. reduction in the number of trainable parameters), and MMAC. While L1, L2, and funnel regularizations have been employed in prior research for model compression during the post-training phases \cite{chu2021low}, we have instead incorporated them into our proposed dynamic compression framework during the training process for a more equitable comparison using $\lambda_\text{reg}=0.1$ and $\epsilon=0.001$ as was done in \cite{chu2021low}. The factorized method without any regularization is our full-rank factorized baseline model.

Our observations indicate that the factorized model, when combined with the proposed regularization, achieves an accuracy of $92.16\%$. This exceeds the baseline and all other regularization methods except the factorized method without any regularization. The latter, while yielding the highest accuracy of $92.32\%$ ($0.16\%$ higher than our proposed method), does so at the expense of an increase in trainable parameters. These results suggest that our proposed regularization technique provides a competitive performance, delivering near-optimal accuracy whilst promoting model compression.

\begin{table}[!t]
\caption{Comparison between regularizations using ResNet-20 baseline network on CIFAR-10 dataset. Best results for each evaluation metric are highlighted in bold text.\label{tab:reg}}
\centering
\fontsize{6}{8.5}\selectfont
\begin{tabularx}{\linewidth}{llYYY}
\toprule
\textbf{Method} & \textbf{Regularization} & \textbf{Top-1 (\%)} & \textbf{Compression (\%)} & \textbf{MMAC}\\
\hline
Baseline & & $90.98$ & $0.00$ & $\mathbf{41.01}$\\
Factorized & & $\mathbf{92.32}$ & $-0.67$ & $43.20$\\
Factorized & L1 & $92.11$ & $0.19$ & $43.18$\\
Factorized & L2 & $92.08$ & $-0.65$ & $43.20$\\
Factorized & Funnel \cite{chu2021low} & $91.29$ & $-0.62$ & $43.20$\\
Factorized & Proposed & $92.16$ & $\mathbf{5.79}$ & $43.04$\\
\bottomrule
\end{tabularx}
\end{table}

\begin{figure*}[t!]
    \centering
    \includegraphics[width=1\linewidth, trim={0cm 4.5cm 0cm 4.5cm},clip]{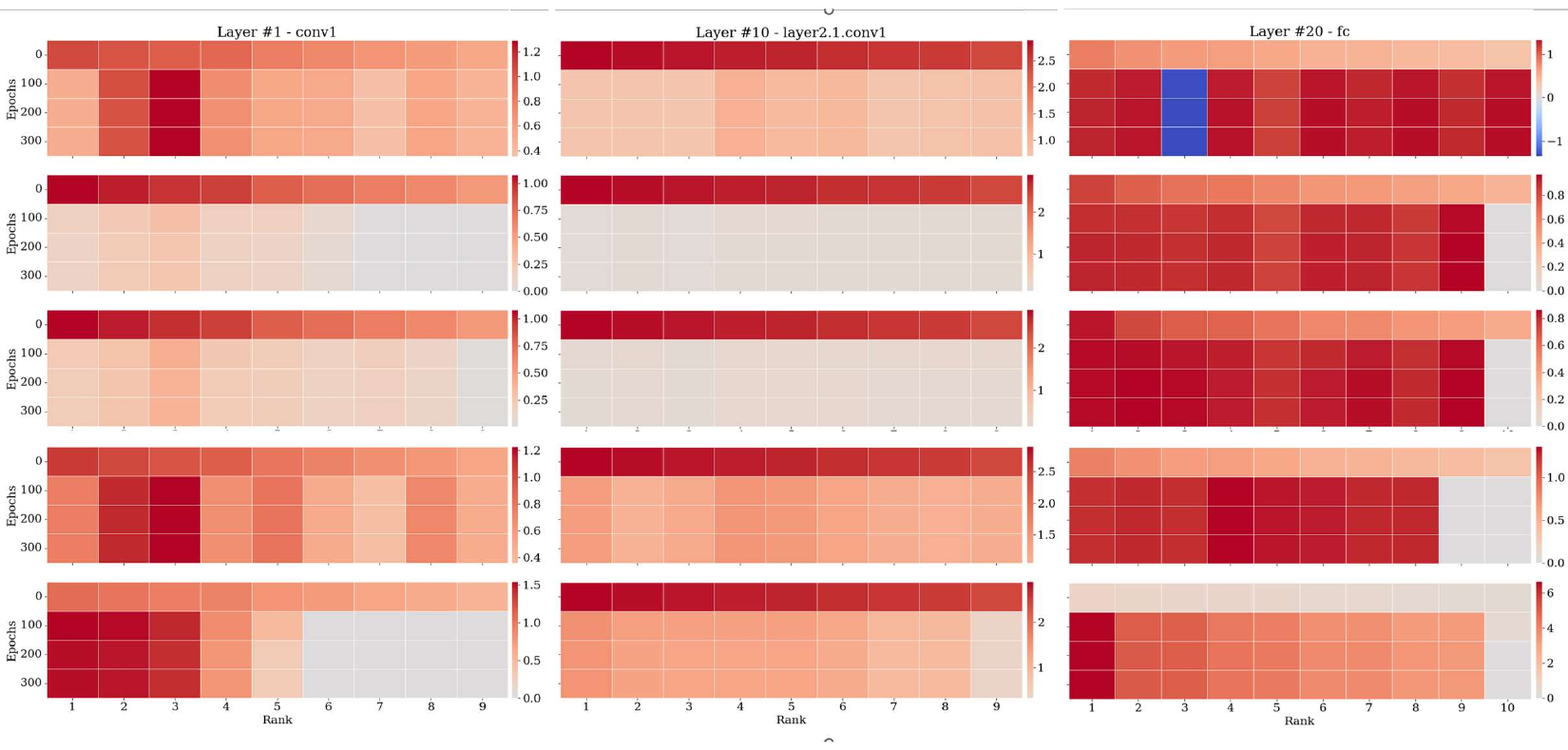}
    \caption{Comparison of rank variation (x-axis) across training epochs (primary y-axis) with a color bar (secondary y-axis) representing the values of singular terms examined at three distinct layers of a ResNet-20 network (from left to right: the initial ($\#1$) layer, the intermediate ($\#10$) layer, and the final ($\#20$) layer) trained on the CIFAR-10 dataset under diverse regularization conditions (from top to bottom: no regularization, L1, L2, funnel, and our proposed regularization).}
    \label{diag_values}
\end{figure*}

Fig. \ref{diag_values} provides additional evidence substantiating our findings. This figure contrasts the rank variation (x-axis) across training epochs (primary y-axis) with a color bar (secondary y-axis) representing the intensity of singular values. We examine this at three distinct layers of the ResNet-20 network: the initial ($\#1$) layer, the intermediate ($\#10$) layer, and the final ($\#20$) layer. We also study the network performance under diverse regularization conditions: no regularization, L1, L2, funnel, and our proposed regularization. All tests are conducted on the CIFAR-10 dataset, with configurations initialized by SVD at epoch $0$.

The first row illustrates a factorized model without any regularization, which departs from the SVD condition during training and exhibits random value fluctuations across all three layers. In contrast, models implementing L1 and L2 regularizations adhere to a more rigorous protocol, suppressing all values during each parameter update in a manner that could be described as `greedy'. Yet, both regularizations lack a focused suppression scheme beneficial for pruning.

The L1 regularization, the most stringent of all, can lead to over-pruning of trainable parameters and subsequent performance degradation. Therefore, it demands cautious selection of pruning thresholds and scaling weights. Although L2 regularization penalizes large deviations from sparsity, its failure to suppress values beyond the pruning threshold undermines its suitability for the compression process.
Funnel regularization strives for rank reduction through a steep loss slope for minimal values. However, it presumptuously anticipates the presence of small singular values across all layers, thus hindering its effectiveness. 

It should be noted that the unregularized factorized method as well as all the above regularizations deviate from the SVD condition, inducing the learning of correlated features and sub-optimal exploration of redundancies in trainable parameters.
In contrast, our proposed regularization method actively encourages adherence to the SVD condition throughout training, exhibiting well-managed rank variations. This method concentrates these variations, prompting sparsity in the least-valued rightmost values, which are dynamically removed during the training phase itself. The focus on SVD conditions during training fosters the learning of uncorrelated parameters, which in turn allows for an optimal exploration of redundancies in trainable parameters. Notably, our proposed regularization deviates from other methods by employing a pruning threshold in relative terms rather than absolute ones. This approach promotes the removal of less significant parameters based on the relative values of singular values sorted in descending order. Consequently, pruning of such less important parameters results in little to no deterioration in performance.

\begin{table}[!t]
\caption{Comparison of Top-1 accuracy, GMAC, and speed-up across different methods using a ResNet-18 network and the ImageNet dataset. Best results for each evaluation metric are highlighted in bold text. Unavailable accuracy scores are indicated by `-'.
\label{tab:imagenet_compartive}}
\centering
\fontsize{6}{8.5}\selectfont
\begin{tabularx}{\linewidth}{lYYYY}
\toprule
\textbf{Method} & \textbf{Top-1 (\%)} & \textbf{Top-5 (\%)} & \textbf{GMAC} & \textbf{Speed-Up}\\
\hline
Baseline & $69.54$ & $88.92$ & $1.82$ & $1.00 \times$\\
SlimNet \cite{liu2017learning,gao2018dynamic,chu2021low} & $67.76$ & $87.63$ & $1.31$ & $1.39 \times$ \\
LCL \cite{dong2017more} & $65.91$ & $86.63$ & $1.19$ & $1.53 \times$\\
CP-TPM \cite{astrid2017cp,chu2021low} & $67.30$ & $-$ & $1.15$ & $1.58 \times$\\
FPGM \cite{he2019filter} & $67.62$ & $87.83$ & $1.06$ & $1.72 \times$\\
DCP \cite{zhuang2018discrimination, gao2018dynamic} & $67.25$ & $87.54$ & $0.96$ & $1.89 \times$\\
SFP \cite{he2018soft} & $66.39$ & $87.08$ & $1.06$ & $1.72 \times$\\
FBS \cite{gao2018dynamic} & $67.04$ & $87.47$ & $0.92$ & $1.98 \times$\\
CGNN \cite{hua2019channel} & $67.91$ & $87.89$ & $1.13$ & $1.61 \times$\\
MUSCO \cite{gusak2019automated,chu2021low,yin2022batude} & $68.72$ & $88.62$ & $0.75$ & $2.42 \times$\\
TRP \cite{xu2021trp} & $65.93$ & $86.72$ & $0.70$ & $2.60 \times$\\
DSA \cite{ning2020dsa} & $68.43$ & $88.20$ & $1.06$ & $1.72 \times$\\
Stable Low-rank \cite{phan2020stable} & $68.85$ & $88.77$ & $\mathbf{0.59}$ & $\mathbf{3.09 \times}$\\
Funnel \cite{chu2021low} & $68.82$ & $-$ & $0.90$ & $2.02 \times$ \\
BATUDE \cite{yin2022batude} & $-$ & $89.25$ & $0.72$ & $2.52 \times$ \\
Proposed & $\mathbf{70.08}$ & $\mathbf{89.62}$ & $1.85$ & $0.98 \times$ \\
\bottomrule
\end{tabularx}
\end{table}

\subsubsection{Performance Analysis on ImageNet Dataset}

In Table \ref{tab:imagenet_compartive}, we compare the proposed method with the baseline and various other methods, using the ResNet-18 network and ImageNet dataset. We specifically evaluate the Top-1 and Top-5 accuracy, GMAC, and the computational speed-up relative to the baseline method. Interestingly, our proposed method achieves the highest Top-1 accuracy of $70.08\%$, making it the only method to exceed the baseline performance in terms of Top-1 accuracy. Although our method does not achieve the lowest GMAC or the highest speed-up, it remains computationally similar to the baseline with a speed-up factor of $0.98$. One particular noteworthy observation is that our proposed method is unique in its ability to dynamically determine the factorization rank per layer in an end-to-end trainable manner, based on the training dataset. Furthermore, it accomplishes model compression during training, thereby avoiding the post-training operation utilized by other comparative methods. This feature results in significant savings in terms of post-training rank determination and processing times. Additionally, a closer analysis of the table reveals the delicate balance between speed-up and Top-1 accuracy. Methods with higher speed-up factors, such as FBS \cite{gao2018dynamic} and funnel \cite{chu2021low}, do not necessarily guarantee superior Top-1 accuracy. This result highlights the effectiveness of our proposed method, which provides the highest Top-1 accuracy while maintaining a computational speed-up nearly identical to the baseline. The proposed method's performance underscores the advantage of its novel, end-to-end trainable approach and the benefits of dynamic compression during the training phase.

\section{Discussion}
\label{sec-discussion}

Our study presents a novel dynamic CNN compression training approach, factorization reshaping, and regularization techniques that have demonstrated exceptional performance in terms of Top-1 accuracy, Top-5 accuracy, model compression, and computational speed-up. The primary focus of the proposed regularizations is to promote SVD condition during training that ensures the learning of uncorrelated parameters. Consequently, it encourages optimal exploration of redundancies in trainable parameters and fosters better generalization. By concentrating rank variations and promoting focused sparsity, our method allows for dynamic pruning of less significant parameters during the training phase. It is distinct from traditional pruning techniques in that it uses a relative threshold based on the sorted singular values instead of an absolute threshold.

Interestingly, this approach results in minimal performance degradation, if any. An essential element of our proposed method is its ability to dynamically determine the factorization rank per layer in an end-to-end trainable manner. This ability is novel compared to other techniques and contributes to significant savings in post-training rank determination and processing times. The contrast between our method and others in terms of computational speed-up provides valuable insights into the trade-off between model efficiency and accuracy. Despite not achieving the highest speed-up, our method ensured a near-baseline computational speed while posting the highest Top-1 accuracy. This delicate balance is a critical factor for practical deployments where computational resources may be limited, but high accuracy is necessary.

However, we recognize the potential trade-offs in our study. The formation of factorized convolutional filters from SVD factors is the main source of additional computational complexity and relatively less computational speed-up. This aspect warrants further investigation and exploration to reduce computational overhead.

\section{Conclusions}
\label{sec-conclusions}

In this paper, we introduced a novel training method that compresses a CNN via DPRP, utilizing an innovative reshaping technique for SVD factorization alongside our proposed regularization techniques. Our method demonstrated superior performance across several key measures such as Top-1 accuracy, Top-5 accuracy, and model compression with competitive computational speeds. The regularization techniques presented a compelling approach to model compression during training via dynamic rank reduction while maintaining high performance in classification tasks. The success of the proposed approach lies in its focus on promoting the SVD condition during training, which facilitates the learning of uncorrelated parameters and dynamic pruning of less significant parameters.
Our findings underscore the importance of carefully balancing model accuracy, network compression, and computational speed-up. Even though achieving the highest computational speed-up is a common objective, our research highlighted the crucial nature of preserving or even improving model accuracy amidst network compression for real-world applications.

Looking forward, there are several avenues to expand our research. Exploring the applicability and performance of our method with different types of neural network architectures, such as transformers or recurrent networks, as well as tasks beyond image classification, like object detection and image segmentation, is a promising direction. Further investigation into determining different hyperparameters dynamically during training could potentially enhance our technique accuracy and compression further. These exciting prospects suggest that our work lays a firm foundation for future research on model compression via dynamic rank determination.






\bibliographystyle{IEEEtran}
\small
\bibliography{main_arXiv}

\begin{thebibliography}{10}
\providecommand{\url}[1]{#1}
\csname url@samestyle\endcsname
\providecommand{\newblock}{\relax}
\providecommand{\bibinfo}[2]{#2}
\providecommand{\BIBentrySTDinterwordspacing}{\spaceskip=0pt\relax}
\providecommand{\BIBentryALTinterwordstretchfactor}{4}
\providecommand{\BIBentryALTinterwordspacing}{\spaceskip=\fontdimen2\font plus
\BIBentryALTinterwordstretchfactor\fontdimen3\font minus \fontdimen4\font\relax}
\providecommand{\BIBforeignlanguage}[2]{{%
\expandafter\ifx\csname l@#1\endcsname\relax
\typeout{** WARNING: IEEEtran.bst: No hyphenation pattern has been}%
\typeout{** loaded for the language `#1'. Using the pattern for}%
\typeout{** the default language instead.}%
\else
\language=\csname l@#1\endcsname
\fi
#2}}
\providecommand{\BIBdecl}{\relax}
\BIBdecl

\bibitem{simonyan2014very}
K.~Simonyan and A.~Zisserman, ``Very deep convolutional networks for large-scale image recognition,'' \emph{arXiv preprint arXiv:1409.1556}, 2014.

\bibitem{he2016deep}
K.~He, X.~Zhang, S.~Ren, and J.~Sun, ``Deep residual learning for image recognition,'' in \emph{Proceedings of the IEEE conference on computer vision and pattern recognition}, 2016, pp. 770--778.

\bibitem{krizhevsky2017imagenet}
A.~Krizhevsky, I.~Sutskever, and G.~E. Hinton, ``Image{N}et classification with deep convolutional neural networks,'' \emph{Communications of the ACM}, vol.~60, no.~6, pp. 84--90, 2017.

\bibitem{dhanaraj2020vehicle}
M.~Dhanaraj, M.~Sharma, T.~Sarkar, S.~Karnam, D.~G. Chachlakis, R.~Ptucha, P.~P. Markopoulos, and E.~Saber, ``Vehicle detection from multi-modal aerial imagery using {YOLO}v3 with mid-level fusion,'' in \emph{Big data II: learning, analytics, and applications}, vol. 11395.\hskip 1em plus 0.5em minus 0.4em\relax SPIE, 2020, pp. 22--32.

\bibitem{sharma2020yolors}
M.~Sharma, M.~Dhanaraj, S.~Karnam, D.~G. Chachlakis, R.~Ptucha, P.~P. Markopoulos, and E.~Saber, ``Y{OLO}rs: Object detection in multimodal remote sensing imagery,'' \emph{IEEE Journal of Selected Topics in Applied Earth Observations and Remote Sensing}, vol.~14, pp. 1497--1508, 2020.

\bibitem{sharma2021yolors}
M.~Sharma, P.~P. Markopoulos, and E.~Saber, ``Y{OLO}rs-lite: A lightweight {CNN} for real-time object detection in remote-sensing,'' in \emph{2021 IEEE International Geoscience and Remote Sensing Symposium IGARSS}.\hskip 1em plus 0.5em minus 0.4em\relax IEEE, 2021, pp. 2604--2607.

\bibitem{singh2023multimodal}
S.~Singh, M.~Sharma, J.~Heard, J.~D. Lew, E.~Saber, and P.~P. Markopoulos, ``Multimodal aerial view object classification with disjoint unimodal feature extraction and fully-connected-layer fusion,'' in \emph{Big Data V: Learning, Analytics, and Applications}, vol. 12522.\hskip 1em plus 0.5em minus 0.4em\relax SPIE, 2023, p. 1252206.

\bibitem{litjens2017survey}
G.~Litjens, T.~Kooi, B.~E. Bejnordi, A.~A.~A. Setio, F.~Ciompi, M.~Ghafoorian, J.~A. Van Der~Laak, B.~Van~Ginneken, and C.~I. S{\'a}nchez, ``A survey on deep learning in medical image analysis,'' \emph{Medical image analysis}, vol.~42, pp. 60--88, 2017.

\bibitem{wen2022deep}
L.-H. Wen and K.-H. Jo, ``Deep learning-based perception systems for autonomous driving: A comprehensive survey,'' \emph{Neurocomputing}, 2022.

\bibitem{denil2013predicting}
M.~Denil, B.~Shakibi, L.~Dinh, M.~Ranzato, and N.~De~Freitas, ``Predicting parameters in deep learning,'' \emph{Advances in neural information processing systems}, vol.~26, 2013.

\bibitem{hinton2015distilling}
G.~Hinton, O.~Vinyals, and J.~Dean, ``Distilling the knowledge in a neural network,'' \emph{arXiv preprint arXiv:1503.02531}, 2015.

\bibitem{zagoruyko2016paying}
S.~Zagoruyko and N.~Komodakis, ``Paying more attention to attention: Improving the performance of convolutional neural networks via attention transfer,'' in \emph{International Conference on Learning Representations}, 2016.

\bibitem{yim2017gift}
J.~Yim, D.~Joo, J.~Bae, and J.~Kim, ``A gift from knowledge distillation: Fast optimization, network minimization and transfer learning,'' in \emph{Proceedings of the IEEE conference on computer vision and pattern recognition}, 2017, pp. 4133--4141.

\bibitem{ahn2019variational}
S.~Ahn, S.~X. Hu, A.~Damianou, N.~D. Lawrence, and Z.~Dai, ``Variational information distillation for knowledge transfer,'' in \emph{Proceedings of the IEEE/CVF Conference on Computer Vision and Pattern Recognition}, 2019, pp. 9163--9171.

\bibitem{park2019relational}
W.~Park, D.~Kim, Y.~Lu, and M.~Cho, ``Relational knowledge distillation,'' in \emph{Proceedings of the IEEE/CVF Conference on Computer Vision and Pattern Recognition}, 2019, pp. 3967--3976.

\bibitem{gong2014compressing}
Y.~Gong, L.~Liu, M.~Yang, and L.~Bourdev, ``Compressing deep convolutional networks using vector quantization,'' \emph{arXiv preprint arXiv:1412.6115}, 2014.

\bibitem{han2015deep}
S.~Han, H.~Mao, and W.~J. Dally, ``Deep compression: Compressing deep neural networks with pruning, trained quantization and huffman coding,'' \emph{arXiv preprint arXiv:1510.00149}, 2015.

\bibitem{courbariaux2016binarized}
M.~Courbariaux, I.~Hubara, D.~Soudry, R.~El-Yaniv, and Y.~Bengio, ``Binarized neural networks: Training deep neural networks with weights and activations constrained to+ 1 or-1,'' \emph{arXiv preprint arXiv:1602.02830}, 2016.

\bibitem{lin2016fixed}
D.~Lin, S.~Talathi, and S.~Annapureddy, ``Fixed point quantization of deep convolutional networks,'' in \emph{International conference on machine learning}.\hskip 1em plus 0.5em minus 0.4em\relax PMLR, 2016, pp. 2849--2858.

\bibitem{zhouincremental}
A.~Zhou, A.~Yao, Y.~Guo, L.~Xu, and Y.~Chen, ``Incremental network quantization: Towards lossless {CNN}s with low-precision weights,'' in \emph{International Conference on Learning Representations}.

\bibitem{molchanov2016pruning}
P.~Molchanov, S.~Tyree, T.~Karras, T.~Aila, and J.~Kautz, ``Pruning convolutional neural networks for resource efficient inference,'' in \emph{International Conference on Learning Representations}, 2016.

\bibitem{lipruning}
H.~Li, A.~Kadav, I.~Durdanovic, H.~Samet, and H.~P. Graf, ``Pruning filters for efficient convnets,'' in \emph{International Conference on Learning Representations}.

\bibitem{iandola2016squeezenet}
F.~N. Iandola, S.~Han, M.~W. Moskewicz, K.~Ashraf, W.~J. Dally, and K.~Keutzer, ``Squeeze{N}et: Alex{N}et-level accuracy with 50x fewer parameters and \textless0.5 mb model size,'' 2016.

\bibitem{chollet2017xception}
F.~Chollet, ``Xception: Deep learning with depthwise separable convolutions,'' in \emph{Proceedings of the IEEE conference on computer vision and pattern recognition}, 2017, pp. 1251--1258.

\bibitem{howard2017mobilenets}
A.~G. Howard, M.~Zhu, B.~Chen, D.~Kalenichenko, W.~Wang, T.~Weyand, M.~Andreetto, and H.~Adam, ``Mobile{N}ets: Efficient convolutional neural networks for mobile vision applications,'' \emph{arXiv preprint arXiv:1704.04861}, 2017.

\bibitem{sharma2021convolutional}
M.~Sharma, P.~P. Markopoulos, E.~Saber, M.~S. Asif, and A.~Prater-Bennette, ``Convolutional auto-encoder with tensor-train factorization,'' in \emph{Proceedings of the IEEE/CVF international conference on computer vision}, 2021, pp. 198--206.

\bibitem{hyder2022incremental}
R.~Hyder, K.~Shao, B.~Hou, P.~Markopoulos, A.~Prater-Bennette, and M.~S. Asif, ``Incremental task learning with incremental rank updates,'' in \emph{European Conference on Computer Vision}.\hskip 1em plus 0.5em minus 0.4em\relax Springer, 2022, pp. 566--582.

\bibitem{denton2014exploiting}
E.~L. Denton, W.~Zaremba, J.~Bruna, Y.~LeCun, and R.~Fergus, ``Exploiting linear structure within convolutional networks for efficient evaluation,'' \emph{Advances in neural information processing systems}, vol.~27, 2014.

\bibitem{cp}
V.~Lebedev, Y.~Ganin, M.~Rakhuba, I.~Oseledets, and V.~Lempitsky, ``Speeding-up convolutional neural networks using fine-tuned {CP}-decomposition,'' \emph{arXiv preprint arXiv:1412.6553}, 2014.

\bibitem{tuc_conv}
Y.-D. Kim, E.~Park, S.~Yoo, T.~Choi, L.~Yang, and D.~Shin, ``Compression of deep convolutional neural networks for fast and low power mobile applications,'' \emph{arXiv preprint arXiv:1511.06530}, 2015.

\bibitem{ult_tensor}
T.~Garipov, D.~Podoprikhin, A.~Novikov, and D.~Vetrov, ``Ultimate tensorization: compressing convolutional and {FC} layers alike,'' \emph{arXiv preprint arXiv:1611.03214}, 2016.

\bibitem{sainath2013low}
T.~N. Sainath, B.~Kingsbury, V.~Sindhwani, E.~Arisoy, and B.~Ramabhadran, ``Low-rank matrix factorization for deep neural network training with high-dimensional output targets,'' in \emph{2013 IEEE international conference on acoustics, speech and signal processing}.\hskip 1em plus 0.5em minus 0.4em\relax IEEE, 2013, pp. 6655--6659.

\bibitem{chung2019parameter}
H.~Chung, E.~Chung, J.~G. Park, and H.-Y. Jung, ``Parameter reduction for deep neural network based acoustic models using sparsity regularized factorization neurons,'' in \emph{2019 International Joint Conference on Neural Networks (IJCNN)}.\hskip 1em plus 0.5em minus 0.4em\relax IEEE, 2019, pp. 1--5.

\bibitem{xu2021trp}
Y.~Xu, Y.~Li, S.~Zhang, W.~Wen, B.~Wang, Y.~Qi, Y.~Chen, W.~Lin, and H.~Xiong, ``T{RP}: Trained rank pruning for efficient deep neural networks,'' in \emph{Proceedings of the Twenty-Ninth International Conference on International Joint Conferences on Artificial Intelligence}, 2021, pp. 977--983.

\bibitem{chu2021low}
B.-S. Chu and C.-R. Lee, ``Low-rank tensor decomposition for compression of convolutional neural networks using funnel regularization,'' \emph{arXiv preprint arXiv:2112.03690}, 2021.

\bibitem{yin2022batude}
M.~Yin, H.~Phan, X.~Zang, S.~Liao, and B.~Yuan, ``B{ATUDE}: Budget-aware neural network compression based on tucker decomposition,'' in \emph{Proceedings of the AAAI Conference on Artificial Intelligence}, vol.~36, no.~8, 2022, pp. 8874--8882.

\bibitem{liu2017learning}
Z.~Liu, J.~Li, Z.~Shen, G.~Huang, S.~Yan, and C.~Zhang, ``Learning efficient convolutional networks through network slimming,'' in \emph{Proceedings of the IEEE international conference on computer vision}, 2017, pp. 2736--2744.

\bibitem{guo2016dynamic}
Y.~Guo, A.~Yao, and Y.~Chen, ``Dynamic network surgery for efficient {DNN}s,'' \emph{Advances in neural information processing systems}, vol.~29, 2016.

\bibitem{huang2018data}
Z.~Huang and N.~Wang, ``Data-driven sparse structure selection for deep neural networks,'' in \emph{Proceedings of the European conference on computer vision (ECCV)}, 2018, pp. 304--320.

\bibitem{kossaifi2017tensor}
J.~Kossaifi, A.~Khanna, Z.~Lipton, T.~Furlanello, and A.~Anandkumar, ``Tensor contraction layers for parsimonious deep nets,'' in \emph{Proceedings of the IEEE Conference on Computer Vision and Pattern Recognition Workshops}, 2017, pp. 26--32.

\bibitem{tran2018improving}
D.~T. Tran, A.~Iosifidis, and M.~Gabbouj, ``Improving efficiency in convolutional neural networks with multilinear filters,'' \emph{Neural Networks}, vol. 105, pp. 328--339, 2018.

\bibitem{kossaifi2019t}
J.~Kossaifi, A.~Bulat, G.~Tzimiropoulos, and M.~Pantic, ``T-net: Parametrizing fully convolutional nets with a single high-order tensor,'' in \emph{Proceedings of the IEEE/CVF conference on computer vision and pattern recognition}, 2019, pp. 7822--7831.

\bibitem{panagakis2024tensor}
Y.~Panagakis, J.~Kossaifi, G.~G. Chrysos, J.~Oldfield, T.~Patti, M.~A. Nicolaou, A.~Anandkumar, and S.~Zafeiriou, ``Chapter 15 - tensor methods in deep learning,'' in \emph{Signal Processing and Machine Learning Theory}.\hskip 1em plus 0.5em minus 0.4em\relax Elsevier, 2024, pp. 1009--1048.

\bibitem{gusak2019automated}
J.~Gusak, M.~Kholiavchenko, E.~Ponomarev, L.~Markeeva, P.~Blagoveschensky, A.~Cichocki, and I.~Oseledets, ``Automated multi-stage compression of neural networks,'' in \emph{Proceedings of the IEEE/CVF International Conference on Computer Vision Workshops}, 2019, pp. 0--0.

\bibitem{yang2020learning}
H.~Yang, M.~Tang, W.~Wen, F.~Yan, D.~Hu, A.~Li, H.~Li, and Y.~Chen, ``Learning low-rank deep neural networks via singular vector orthogonality regularization and singular value sparsification,'' in \emph{Proceedings of the IEEE/CVF conference on computer vision and pattern recognition workshops}, 2020, pp. 678--679.

\bibitem{frankle2018lottery}
J.~Frankle and M.~Carbin, ``The lottery ticket hypothesis: Finding sparse, trainable neural networks,'' in \emph{International Conference on Learning Representations}, 2018.

\bibitem{jaderberg2014speeding}
M.~Jaderberg, A.~Vedaldi, and A.~Zisserman, ``Speeding up convolutional neural networks with low rank expansions,'' \emph{arXiv preprint arXiv:1405.3866}, 2014.

\bibitem{zhang2015accelerating}
X.~Zhang, J.~Zou, K.~He, and J.~Sun, ``Accelerating very deep convolutional networks for classification and detection,'' \emph{IEEE transactions on pattern analysis and machine intelligence}, vol.~38, no.~10, pp. 1943--1955, 2015.

\bibitem{phan2020stable}
A.-H. Phan, K.~Sobolev, K.~Sozykin, D.~Ermilov, J.~Gusak, P.~Tichavsk{\`y}, V.~Glukhov, I.~Oseledets, and A.~Cichocki, ``Stable low-rank tensor decomposition for compression of convolutional neural network,'' in \emph{Computer Vision--ECCV 2020: 16th European Conference, Glasgow, UK, August 23--28, 2020, Proceedings, Part XXIX 16}.\hskip 1em plus 0.5em minus 0.4em\relax Springer, 2020, pp. 522--539.

\bibitem{kim2015compression}
Y.-D. Kim, E.~Park, S.~Yoo, T.~Choi, L.~Yang, and D.~Shin, ``Compression of deep convolutional neural networks for fast and low power mobile applications,'' \emph{arXiv preprint arXiv:1511.06530}, 2015.

\bibitem{astrid2017cp}
M.~Astrid and S.-I. Lee, ``C{P}-decomposition with tensor power method for convolutional neural networks compression,'' in \emph{2017 IEEE International Conference on Big Data and Smart Computing (BigComp)}.\hskip 1em plus 0.5em minus 0.4em\relax IEEE, 2017, pp. 115--118.

\bibitem{li2021heuristic}
N.~Li, Y.~Pan, Y.~Chen, Z.~Ding, D.~Zhao, and Z.~Xu, ``Heuristic rank selection with progressively searching tensor ring network,'' \emph{Complex \& Intelligent Systems}, pp. 1--15, 2021.

\bibitem{zangrando2023rank}
E.~Zangrando, S.~Schotth{\"o}fer, G.~Ceruti, J.~Kusch, and F.~Tudisco, ``Rank-adaptive spectral pruning of convolutional layers during training,'' \emph{arXiv preprint arXiv:2305.19059}, 2023.

\bibitem{golub2013matrix}
G.~H. Golub and C.~F. Van~Loan, \emph{Matrix computations}.\hskip 1em plus 0.5em minus 0.4em\relax JHU press, 2013.

\bibitem{russakovsky2015imagenet}
O.~Russakovsky, J.~Deng, H.~Su, J.~Krause, S.~Satheesh, S.~Ma, Z.~Huang, A.~Karpathy, A.~Khosla, M.~Bernstein \emph{et~al.}, ``Image{N}et large scale visual recognition challenge,'' \emph{International journal of computer vision}, vol. 115, pp. 211--252, 2015.

\bibitem{krizhevsky2012imagenet}
A.~Krizhevsky, I.~Sutskever, and G.~E. Hinton, ``Image{N}et classification with deep convolutional neural networks,'' \emph{Advances in neural information processing systems}, vol.~25, 2012.

\bibitem{ning2020dsa}
X.~Ning, T.~Zhao, W.~Li, P.~Lei, Y.~Wang, and H.~Yang, ``D{SA}: More efficient budgeted pruning via differentiable sparsity allocation,'' in \emph{European Conference on Computer Vision}.\hskip 1em plus 0.5em minus 0.4em\relax Springer, 2020, pp. 592--607.

\bibitem{gao2018dynamic}
X.~Gao, Y.~Zhao, {\L}.~Dudziak, R.~Mullins, and C.-z. Xu, ``Dynamic channel pruning: Feature boosting and suppression,'' in \emph{International Conference on Learning Representations}, 2018.

\bibitem{dong2017more}
X.~Dong, J.~Huang, Y.~Yang, and S.~Yan, ``More is less: A more complicated network with less inference complexity,'' in \emph{Proceedings of the IEEE conference on computer vision and pattern recognition}, 2017, pp. 5840--5848.

\bibitem{he2019filter}
Y.~He, P.~Liu, Z.~Wang, Z.~Hu, and Y.~Yang, ``Filter pruning via geometric median for deep convolutional neural networks acceleration,'' in \emph{Proceedings of the IEEE/CVF conference on computer vision and pattern recognition}, 2019, pp. 4340--4349.

\bibitem{zhuang2018discrimination}
Z.~Zhuang, M.~Tan, B.~Zhuang, J.~Liu, Y.~Guo, Q.~Wu, J.~Huang, and J.~Zhu, ``Discrimination-aware channel pruning for deep neural networks,'' \emph{Advances in neural information processing systems}, vol.~31, 2018.

\bibitem{he2018soft}
Y.~He, G.~Kang, X.~Dong, Y.~Fu, and Y.~Yang, ``Soft filter pruning for accelerating deep convolutional neural networks,'' in \emph{Proceedings of the 27th International Joint Conference on Artificial Intelligence}, 2018, pp. 2234--2240.

\bibitem{hua2019channel}
W.~Hua, Y.~Zhou, C.~M. De~Sa, Z.~Zhang, and G.~E. Suh, ``Channel gating neural networks,'' \emph{Advances in Neural Information Processing Systems}, vol.~32, 2019.

\end{thebibliography}

\end{document}